\pdfoutput=1
\documentclass[11pt]{article}
\usepackage{acl2010}
\makeatletter
\newcommand{\@BIBLABEL}{\@emptybiblabel}
\newcommand{\@emptybiblabel}[1]{}
\makeatother
\usepackage{times}
\usepackage{url}
\usepackage{latexsym}
\usepackage{graphicx}
\usepackage{fancyhdr} 
\usepackage{hyperref}

\title{Bucking the Trend: Large-Scale Cost-Focused Active Learning for Statistical Machine Translation}

\author{Michael Bloodgood\\
Human Language Technology\\
Center of Excellence\\
Johns Hopkins University\\
Baltimore, MD 21211\\
{\tt bloodgood@jhu.edu}  \And
Chris Callison-Burch\\
Center for Language and\\
Speech Processing\\
Johns Hopkins University\\
Baltimore, MD 21211\\
{\tt  ccb@cs.jhu.edu}}

\date{}

\begin{document}

\thispagestyle{fancy}

\maketitle
\begin{abstract}
  We explore how to improve machine translation systems by adding more translation data in situations where we already have substantial resources.
  The main challenge is how to buck the trend of diminishing returns that is commonly encountered. 
  We present an active learning-style data solicitation algorithm to meet this challenge. 
  We test it, gathering annotations via Amazon Mechanical Turk, and find that we get an order of magnitude increase in performance rates of improvement. 
\end{abstract}

\section{Introduction} \label{introduction}

\begin{figure}[t]
\begin{center}
\includegraphics[width=.5\textwidth]{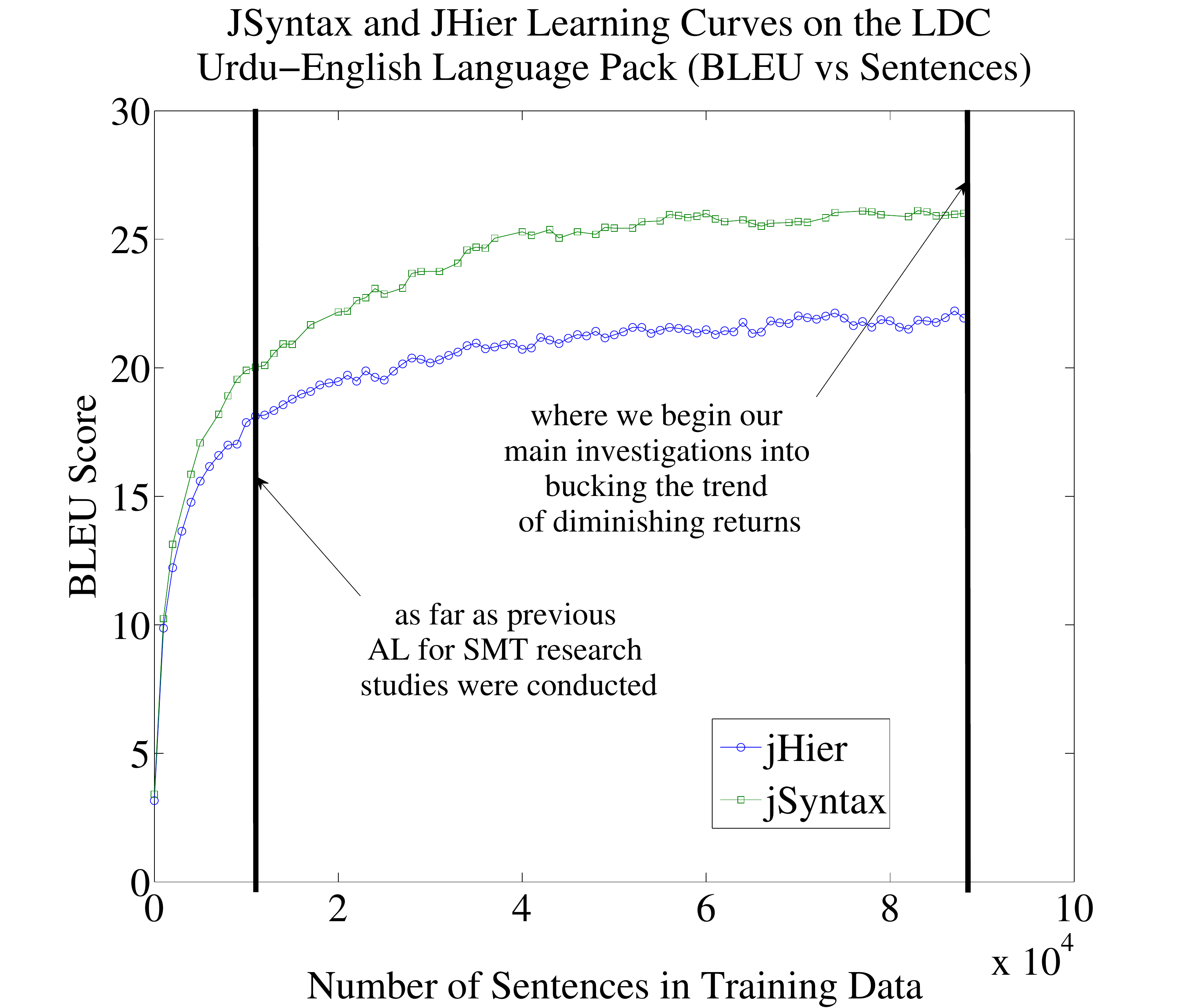}
\caption{\label{f:flatteningCurves} 
Syntax-based and Hierarchical Phrase-Based MT systems' learning curves on the LDC Urdu-English language pack. The x-axis measures the number of sentence pairs in the training data. The y-axis measures BLEU score. 
Note the diminishing returns as more data is added. Also note how relatively early on in the process previous studies were terminated. In contrast, the focus of our 
main experiments doesn't even begin until much higher performance has already been achieved with a period of diminishing returns firmly established.}
\end{center}
\end{figure}

Figure~\ref{f:flatteningCurves} shows the learning curves for two state of the art statistical machine translation (SMT) systems for Urdu-English translation.
Observe how the learning curves rise rapidly at first but then a trend of diminishing returns occurs: put simply, the curves flatten. 

This paper investigates whether we can buck the trend of diminishing returns, and if so, how we can do it effectively. 
Active learning (AL) has been applied to SMT recently \cite{haffari2009a,haffari2009b} but they were interested in starting with 
a tiny seed set of data, and they stopped their investigations after only adding a relatively tiny amount of data as depicted in Figure~\ref{f:flatteningCurves}.

In contrast, we are interested in applying AL when a large amount of data already exists as is the case for many important lanuage pairs. 
We develop an AL algorithm that focuses on keeping annotation costs (measured by time in seconds) low. 
It succeeds in doing this by only soliciting translations for parts of sentences. We show that this gets a savings in human annotation time above and beyond what the
reduction in \# words annotated would have indicated by a factor of about three and speculate as to why.

We conduct experiments for Urdu-English translation, gathering annotations via Amazon Mechanical Turk (MTurk) and show that we can indeed buck the trend of diminishing
returns, achieving an order of magnitude increase in the rate of improvement in performance. 

Section~\ref{related} discusses related work; Section~\ref{simulations} discusses preliminary experiments that show the guiding principles behind the algorithm we
use; Section~\ref{method} explains our method for soliciting new translation data; Section~\ref{experiments} presents our main results; and Section~\ref{conclusions}
concludes.  

\section{Related Work} \label{related}

Active learning has been shown to be effective for improving NLP systems and reducing annotation burdens for a number of NLP tasks (see, e.g.,
\cite{hwa2000,sassano2002,bloodgood2008,bloodgood2009a,mairesse2010,vickrey2010}). 
The current paper is most highly related to previous work falling into three main areas: use of AL when large corpora already exist;
cost-focused AL; and AL for SMT. 

In a sense, the work of \newcite{banko2001} is closely related to ours. 
Though their focus is mainly on investigating the performance of learning methods on giant corpora many orders of magnitude larger than previously used, they do 
lay out how AL might be useful to apply to acquire data to augment a large set cheaply because they recognize the problem of diminishing
returns that we discussed in Section~\ref{introduction}. 

The second area of work that is related to ours is previous work on AL that is cost-conscious. The vast majority of AL research has not focused on accurate cost
accounting and a typical assumption is that each annotatable has equal annotation cost. 
An early exception in the AL for NLP field was the work of \newcite{hwa2000}, which makes a point of using \# of brackets to measure cost for a syntactic analysis
task instead of using \# of sentences. 
Another relatively early work in our field along these lines was the work of \newcite{ngai2000}, which measured actual times of annotation to compare the
efficacy of rule writing versus annotation with AL for the task of BaseNP chunking. 
\newcite{osborne2004} argued for the use of discriminant cost over unit cost for the task of Head Phrase Structure Grammar parse selection.
\newcite{king2004} design a robot that tests gene functions. The robot chooses which experiments to conduct by using AL and takes monetary costs (in pounds sterling)
into account during AL selection and evaluation. 
Unlike our situation for SMT, their costs are all known beforehand because they are simply the cost of materials to conduct the experiments, which are already known
to the robot. 
\newcite{hachey2005} showed that selectively sampled examples for an NER task took longer to annotate and had lower inter-annotator agreement. 
This work is related to ours because it shows that how examples are selected can impact the cost of annotation, an idea we turn around to use for our advantage 
when developing our data selection algorithm. 
\newcite{haertel2008} emphasize measuring costs carefully for AL for POS tagging. They develop a model based on a user study that can estimate the time required for
POS annotating. 
\newcite{kapoor2007} assign costs for AL based on message length for a voicemail classification task. In contrast, we show for SMT that annotation times do not 
scale according to length in words and we show our method can achieve a speedup in annotation time above and beyond what the reduction in words would indicate. 
\newcite{tomanek2009b} measure cost by \# of tokens for an NER task. Their AL method only solicits labels for parts of sentences in the interest of reducing
annotation effort. Along these lines, our method is similar in the respect that we also will only solicit annotation for parts of sentences, 
though we prefer to measure cost with time and we show that time doesn't track with token length for SMT. 

\newcite{haffari2009a}, \newcite{haffari2009b}, and \newcite{ambati2010} investigate AL for SMT. 
There are two major differences between our work and this previous work. 
One is that our intended use cases are very different. 
They deal with the more traditional AL setting of starting from an extremely small set of seed data. 
Also, by SMT standards, they only add a very tiny amount of data during AL. 
All their simulations top out at 10,000 sentences of labeled data and the models learned have relatively low translation quality compared to the state of the art. 

On the other hand, in the current paper, we demonstrate how to apply AL in situations where we already have large corpora. 
Our goal is to buck the trend of diminishing returns and use AL to add data to build some of the highest-performing MT systems in the world while keeping annotation
costs low. 
See Figure~\ref{f:flatteningCurves} from Section~\ref{introduction}, which contrasts where \cite{haffari2009a,haffari2009b} {\em stop} their
investigations with where we {\em begin} our studies. 

The other major difference is that \cite{haffari2009a,haffari2009b} measure annotation cost by \# of sentences. 
In contrast, we bring to light some potential drawbacks of this practice, showing it can lead to different conclusions than if other annotation cost metrics are used,
such as time and money, which are the metrics that we use.

\section{Simulation Experiments} \label{simulations}

Here we report on results of simulation experiments that help to illustrate and motivate the design decisions of the algorithm we present in Section~\ref{method}.
We use the Urdu-English language pack\footnote{LDC Catalog No.: LDC2006E110.} from the Linguistic Data Consortium (LDC), which contains $\approx$ 88000
Urdu-English sentence translation pairs, amounting to $\approx$ 1.7 million Urdu words translated into English. 
All experiments in this paper evaluate on a genre-balanced split of the NIST2008 Urdu-English test set.
In addition, the language pack contains an Urdu-English dictionary consisting of $\approx$ 114000 entries. 
In all the experiments, we use the dictionary at every iteration of training. 
This will make it harder for us to show our methods providing substantial gains since the dictionary will provide a higher base performance
to begin with. 
However, it would be artificial to ignore dictionary resources when they exist. 

We experiment with two translation models: hierarchical phrase-based translation \cite{chiang2007} and syntax augmented translation \cite{zollmann2006}, both of which are 
implemented in the Joshua decoder \cite{li2009}. We hereafter refer to these systems as jHier and jSyntax, respectively. 

We will now present results of experiments with different methods for growing MT training data. 
The results are organized into three areas of investigations:
\begin{enumerate}
\item annotation costs;
\item managing uncertainty; and
\item how to automatically detect when to stop soliciting annotations from a pool of data.
\end{enumerate}

\subsection{Annotation Costs} \label{costs}

We begin our cost investigations with four simple methods for growing MT training data: random, shortest, longest, and {\em VocabGrowth} sentence selection. 
The first three methods are self-explanatory. 
{\em VocabGrowth} (hereafter {\em VG}) selection is modeled after the best methods from previous work \cite{haffari2009a,haffari2009b}, which are based on preferring 
sentences that contain phrases that occur frequently in unlabeled data and infrequently in the so-far labeled data. 
Our {\em VG} method selects sentences for translation that contain n-grams (for n in \{1,2,3,4\}) that do not occur at all in our so-far labeled data. 
We call an n-gram ``covered'' if it occurs at least once in our so-far labeled data. 
{\em VG} has a preference for covering frequent n-grams before covering infrequent n-grams. 
The {\em VG} method is depicted in Figure~\ref{f:vgAlg}.

\begin{figure}
\begin{center}
\begin{tabbing}
{\bf Init:}\=\\
\>Go through all available training\\
\>data (labeled and unlabeled)\\
\>and obtain frequency counts for\\
\>every n-gram (n in $\{1,2,3,4\}$)\\
\>that occurs.\\ 
\>$sortedNGrams \leftarrow$ Sort n-grams by\\ 
\>frequency in descending order.\\
{\bf Loop} \\
until stopping criterion (see Section~\ref{stopping}) is met\\
\>1. $trigger \leftarrow$ Go down $sortedNGrams$ list\\  
\> and find the first n-gram that isn't covered in\\
\> the so far labeled training data.\\
\>2. $selectedSentence \leftarrow$ Find a sentence\\
\> that contains $trigger$. \\
\>3. Remove $selectedSentence$ from unlabeled\\
\> data and add it to labeled training data.\\
{\bf End Loop}\\
\end{tabbing}
\end{center}
\caption{\label{f:vgAlg} The {\em VG} sentence selection algorithm}
\end{figure}

\begin{figure}[t]
\begin{center}
\includegraphics[width=.5\textwidth]{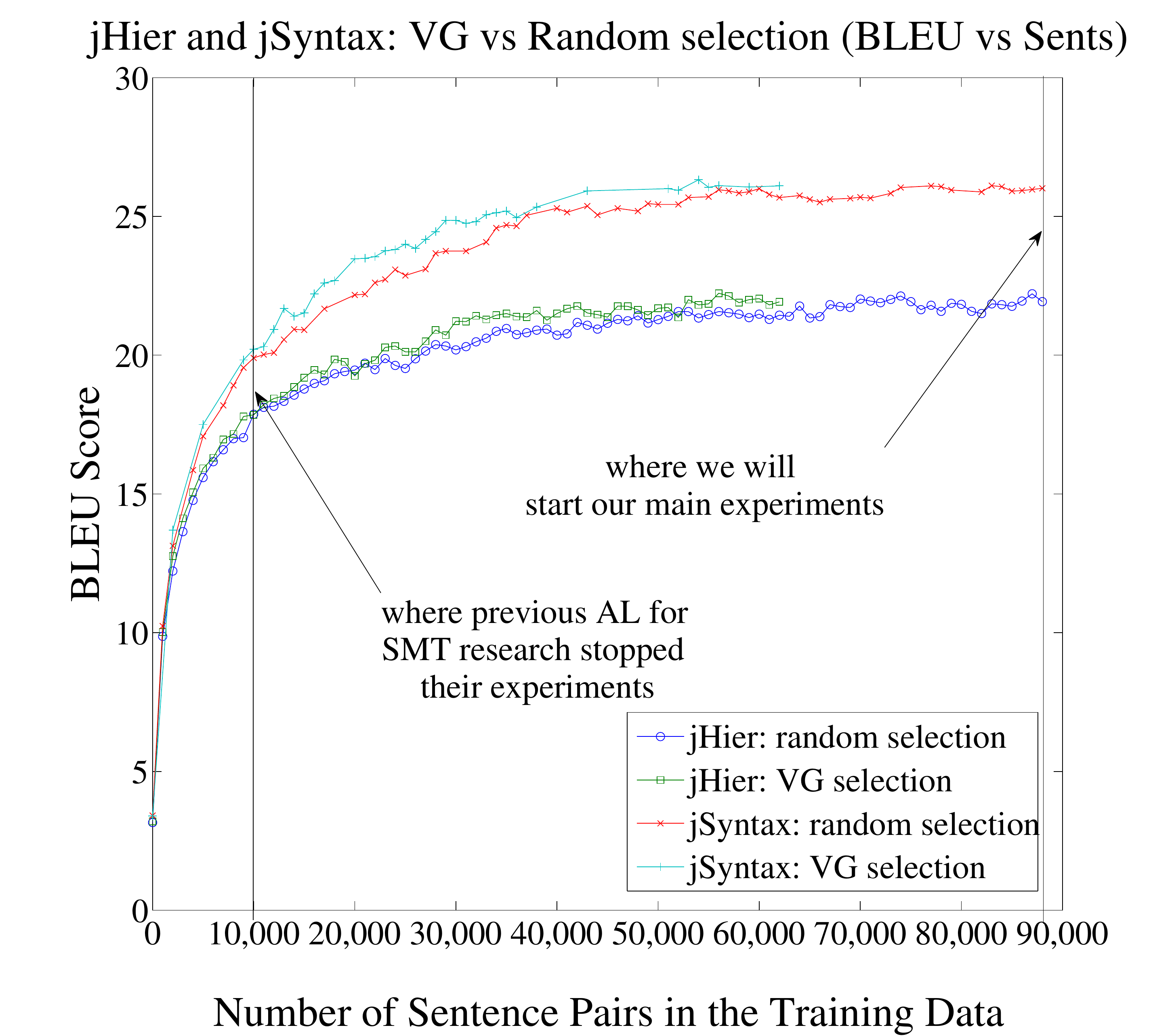}
\caption{\label{f:jAllVGRandSentsWithLines}Random vs {\em VG} selection. The x-axis measures the number of sentence pairs in the training data. The y-axis measures BLEU score.}
\end{center}
\end{figure}

Figure~\ref{f:jAllVGRandSentsWithLines} shows the learning curves for both jHier and jSyntax for {\em VG} selection and random selection.
The y-axis measures BLEU score \cite{papineni2002},which is a fast automatic way of measuring translation quality that has been shown to correlate with human 
judgments and is perhaps the most widely used metric in the MT community. 
The x-axis measures the number of sentence translation pairs in the training data.
The {\em VG} curves are cut off at the point at which the stopping criterion in Section~\ref{stopping} is met.
From Figure~\ref{f:jAllVGRandSentsWithLines} it might appear that {\em VG} selection is better than random selection, achieving higher-performing systems with fewer
translations in the labeled data. 

\begin{figure}[t]
\begin{center}
\includegraphics[width=.5\textwidth]{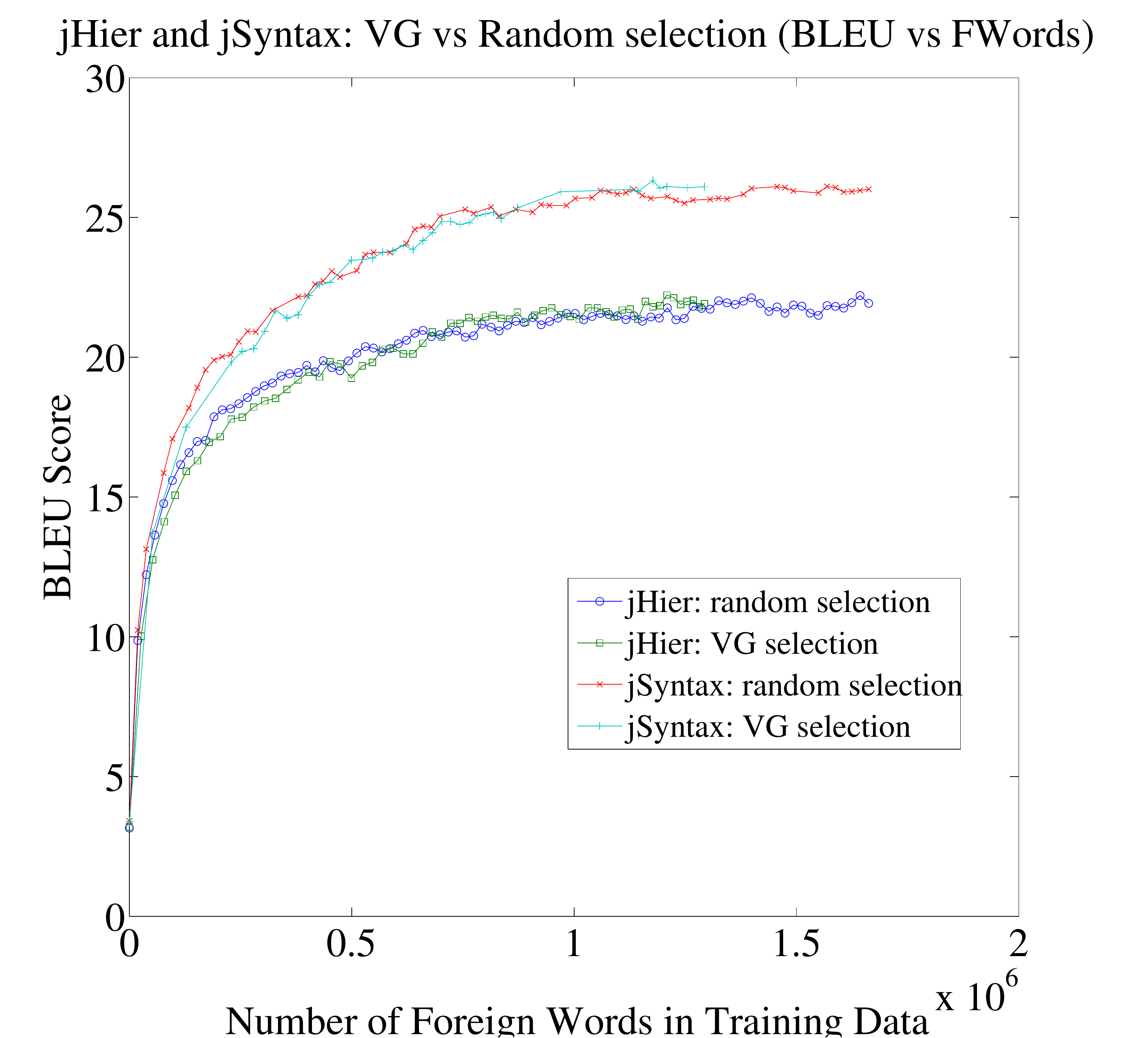}
\caption{\label{f:jAllVGRandWords}Random vs {\em VG} selection. The x-axis measures the number of foreign words in the training data. The y-axis measures BLEU score.}
\end{center}
\end{figure}

However, it is important to take care when measuring annotation costs (especially for relatively complicated tasks such as translation).
Figure~\ref{f:jAllVGRandWords} shows the learning curves for the same systems and selection methods as in Figure~\ref{f:jAllVGRandSentsWithLines} but now the x-axis measures
the number of foreign words in the training data. 
The difference between {\em VG} and random selection now appears smaller. 

\begin{figure}[t]
\begin{center}
\includegraphics[width=.5\textwidth]{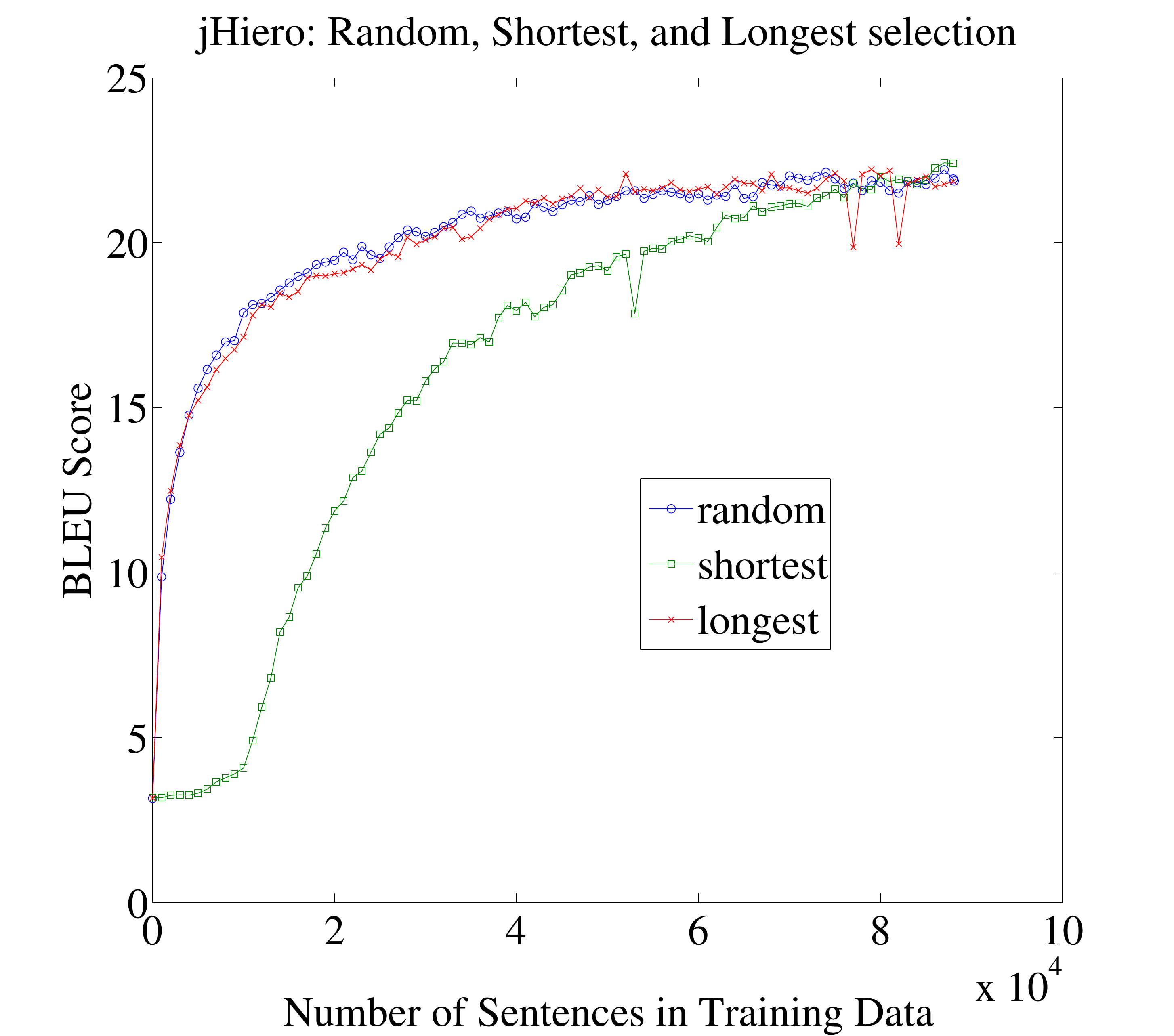}
\caption{\label{f:jHierRandShortLongSents}Random vs Shortest vs Longest selection. The x-axis measures the number of sentence pairs in the training data. The y-axis measures BLEU score.}
\end{center}
\end{figure}

\begin{figure}[t]
\begin{center}
\includegraphics[width=.5\textwidth]{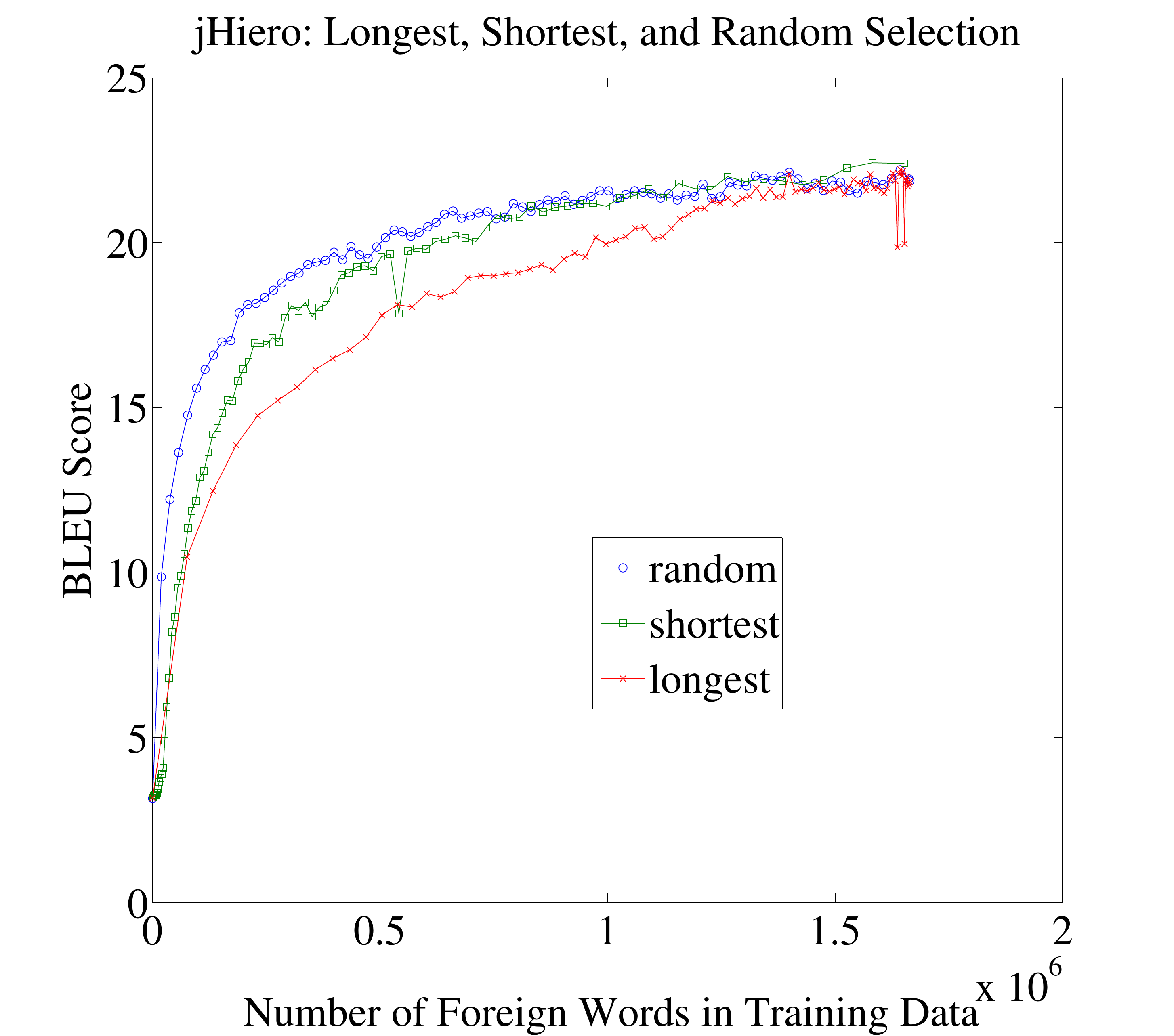}
\caption{\label{f:jHierRandShortLongWords}Random vs Shortest vs Longest selection. The x-axis measures the number of foreign words in the training data. The y-axis measures BLEU score.}
\end{center}
\end{figure}

For an extreme case, to illustrate the ramifications of measuring translation
annotation cost by \# of sentences versus \# of words, consider Figures~\ref{f:jHierRandShortLongSents} and \ref{f:jHierRandShortLongWords}. 
They both show the same three selection methods but Figure~\ref{f:jHierRandShortLongSents} measures the x-axis by \# of sentences and
Figure~\ref{f:jHierRandShortLongWords} measures by \# of words. In Figure~\ref{f:jHierRandShortLongSents}, one would conclude that shortest is a far inferior
selection method to longest but in Figure~\ref{f:jHierRandShortLongWords} one would conclude the opposite. 

Measuring annotation time and cost in dollars are probably the most important measures of annotation cost. 
We can't measure these for the simulated experiments but we will use time (in seconds) and money (in US dollars) as cost measures in Section~\ref{experiments}, which discusses our 
non-simulated AL experiments. 
If \# sentences or \# words track these other more relevant costs in predictable known relationships, then it would suffice to measure \# sentences or \# words
instead. But it's clear that different sentences can have very different annotation time requirements according to how long and complicated they are so we will not use \#
sentences as an annotation cost any more. 
It is not as clear how \# words tracks with annotation time. 
In Section~\ref{experiments} we will present evidence showing that time per word can vary considerably and also show a
method for soliciting annotations that reduces time per word by nearly a factor of three. 

As it is prudent to evaluate using accurate cost accounting, so it is also prudent to develop new AL algorithms that take costs carefully into account. 
Hence, reducing annotation time burdens instead of the \# of sentences translated (which might be quite a different thing) will be a cornerstone of the algorithm we
describe in Section~\ref{method}. 
 
\subsection{Managing Uncertainty} \label{uncertainty}

One of the most successful of all AL methods developed to date is uncertainty sampling and it has been applied successfully many times (e.g.,\cite{lewis1994,tong2002}). 
The intuition is clear: much can be learned (potentially) if there is great uncertainty.  
However, with MT being a relatively complicated task (compared with binary classification, for example), it might be the case that the uncertainty approach has to be
re-considered. 
If words have never occurred in the training data, then uncertainty can be expected to be high. 
But we are concerned that if a sentence is translated for which (almost) no words have been seen in training yet, though uncertainty will be high (which is usually
considered good for AL), the word alignments may be incorrect and then subsequent learning from that translation pair will be severely hampered. 

\begin{figure}[t]
\begin{center}
\includegraphics[width=.5\textwidth]{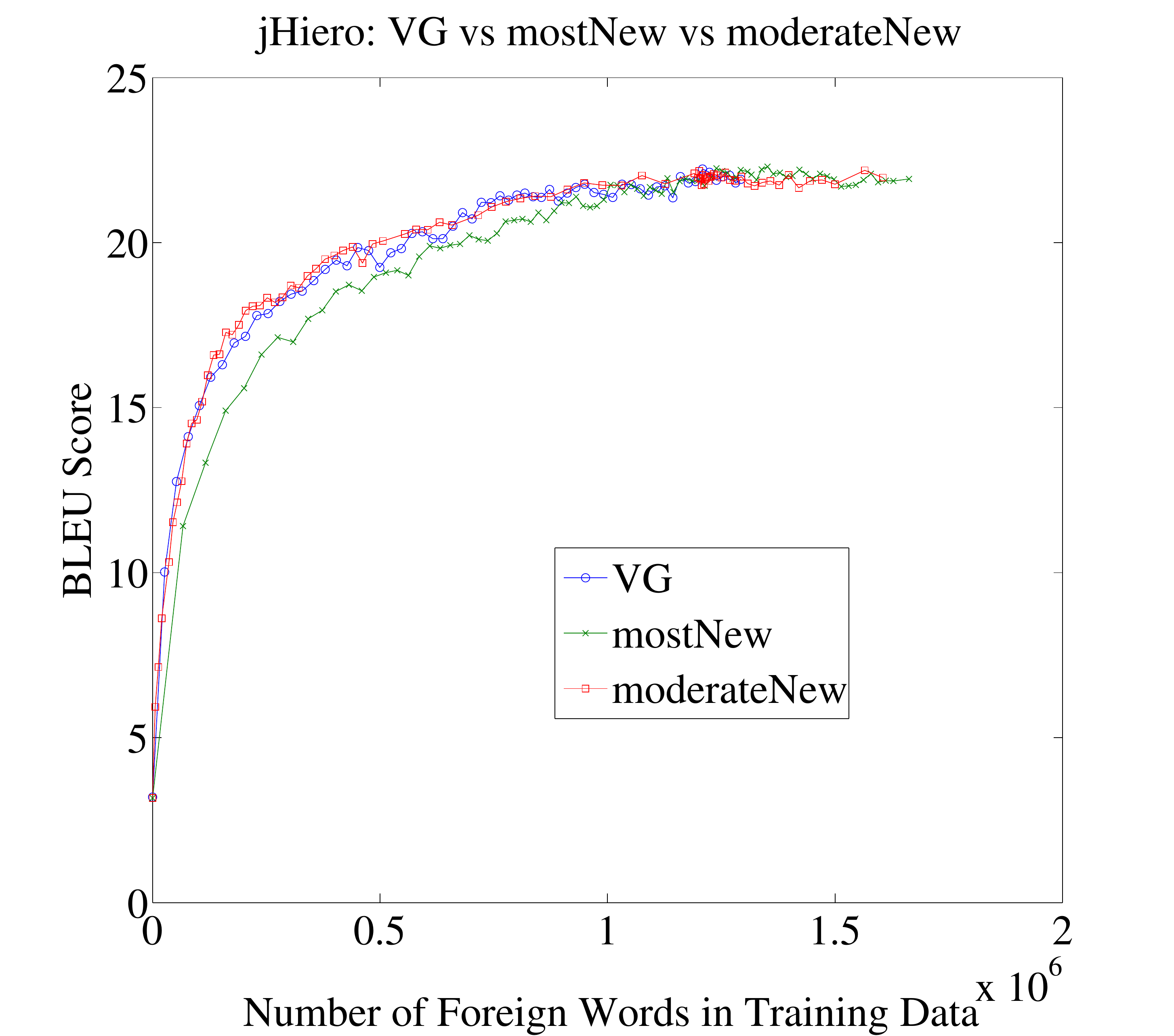}
\caption{\label{f:jHierVGMostNewModerateNewWords}{\em VG} vs MostNew vs ModerateNew selection. The x-axis measures the number of sentence pairs in the training data. The y-axis measures BLEU score.}
\end{center}
\end{figure}

We tested this hypothesis and Figure~\ref{f:jHierVGMostNewModerateNewWords} shows empirical evidence that it is true. 
Along with {\em VG}, two other selection methods' learning curves are charted in Figure~\ref{f:jHierVGMostNewModerateNewWords}: mostNew, which prefers to select
those sentences which have the largest \# of unseen words in them; and moderateNew, which aims to prefer sentences that have a moderate \# of unseen words, preferring
sentences with $\approx$ ten unknown words in them. 
One can see that mostNew underperforms {\em VG}. This could have been due to {\em VG}'s frequency component, which mostNew doesn't have. 
But moderateNew also doesn't have a frequency preference so it is likely that mostNew winds up overwhelming the MT training system, word alignments are
incorrect, and less is learned as a result. In light of this, the algorithm we develop in Section~\ref{method} will be designed to avoid this word alignment danger. 

\subsection{Automatic Stopping} \label{stopping}

The problem of automatically detecting when to stop AL is a substantial one, discussed at length in the literature (e.g.,
\cite{bloodgood2009b,schohn2000,vlachos2008}). 
In our simulation, we stop {\em VG} once all n-grams (n in \{1,2,3,4\}) have been covered. 
Though simple, this stopping criterion seems to work well as can be seen by where the curve for {\em VG} is cut off in Figures~\ref{f:jAllVGRandSentsWithLines} and
\ref{f:jAllVGRandWords}. 
It stops after 1,293,093 words have been translated, with jHier's BLEU=21.92 and jSyntax's BLEU=26.10 at the stopping point.
The ending BLEU scores (with the full corpus annotated) are 21.87 and 26.01 for jHier and jSyntax, respectively. 
So our stopping criterion saves 22.3\% of the annotation (in terms of words) and actually achieves slightly higher BLEU scores than if all the data were used. 
Note: this "less is more" phenomenon has been commonly observed in AL settings (e.g., \cite{bloodgood2009b,schohn2000}). 

\section{Highlighted N-Gram Method} \label{method}

\begin{figure}[t]
\begin{center}
\includegraphics[width=.45\textwidth]{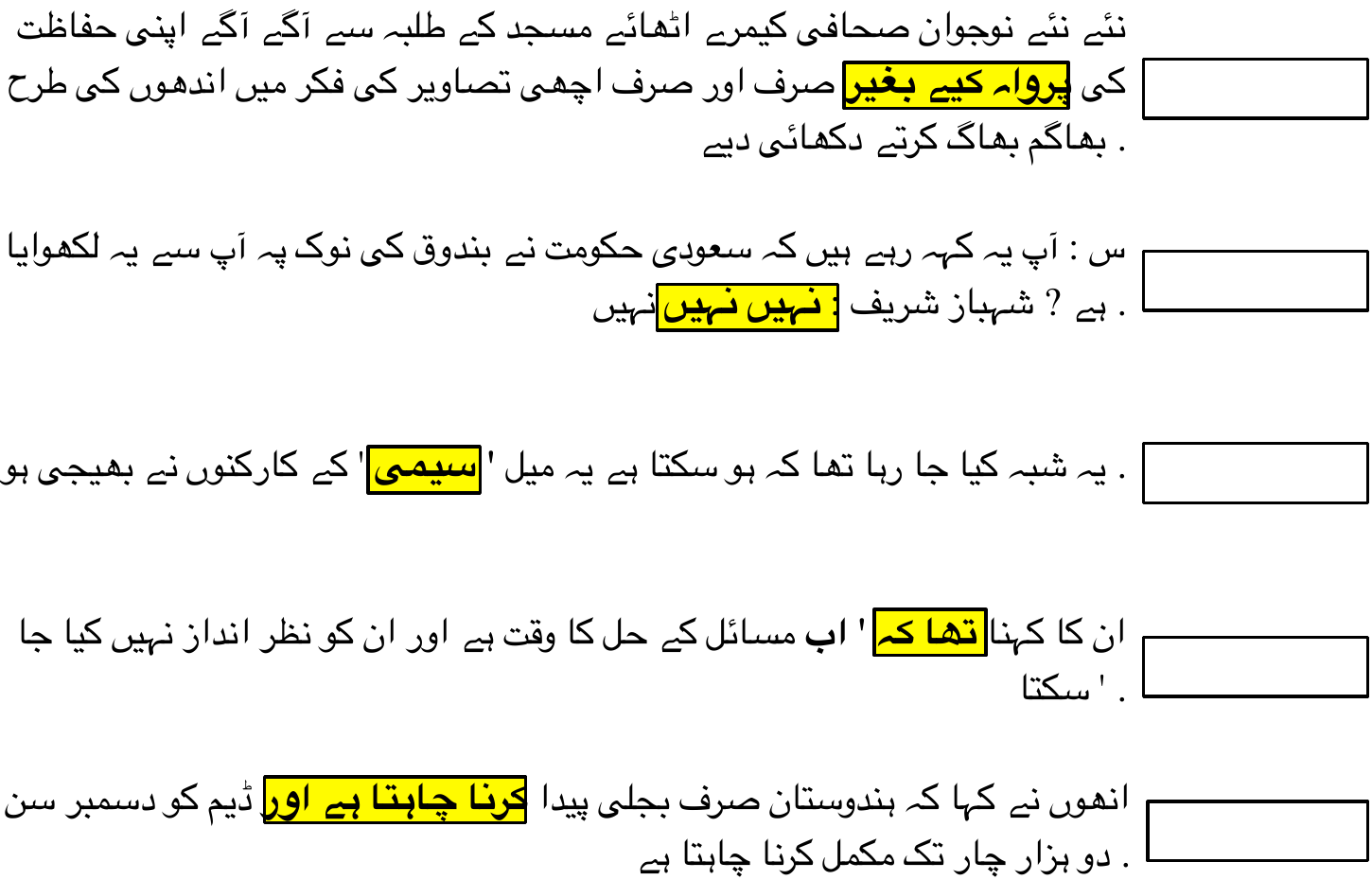}
\caption{\label{f:interface}Screenshot of the interface we used for soliciting translations for triggers.}
\end{center}
\vspace*{-.6cm}
\end{figure}

In this section we describe a method for soliciting human translations that we have applied successfully to improving translation quality in real (not simulated)
conditions. 
We call the method the {\em Highlighted N-Gram} method, or {\em HNG}, for short. 
{\em HNG} solicits translations only for trigger n-grams and not for entire sentences. 
We provide sentential context, highlight the trigger n-gram that we want translated, and ask for a translation of just the highlighted trigger n-gram. 
{\em HNG} asks for translations for triggers in the same order that the triggers are encountered by the algorithm in Figure~\ref{f:vgAlg}. 
A screenshot of our interface is depicted in Figure~\ref{f:interface}. 
The same stopping criterion is used as was used in the last section. When the stopping criterion becomes true, it is time to tap a new unlabeled pool of foreign text,
if available. 

Our motivations for soliciting translations for only parts of sentences are twofold, corresponding to two possible cases. 
Case one is that a translation model learned from the so-far labeled data will be able to translate most of the non-trigger words in the sentence correctly. 
Thus, by asking a human to translate only the trigger words, we avoid wasting human translation effort. 
(We will show in the next section that we even get a much larger speedup above and beyond what the reduction in number of translated words would give us.) 
Case two is that a translation model learned from the so-far labeled data will (in addition to not being able to translate the trigger words correctly) also not be
able to translate most of the non-trigger words correctly. 
One might think then that this would be a great sentence to have translated because the machine can potentially learn a lot from the translation. 
Indeed, one of the overarching themes of AL research is to query examples where uncertainty is greatest. 
But, as we showed evidence for in the last section, for the case of SMT, too much uncertainty could in a sense overwhelm the machine and it might be better to
provide new training data in a more gradual manner. 
A sentence with large \#s of unseen words is likely to get word-aligned incorrectly and then learning from that translation could be hampered. 
By asking for a translation of only the trigger words, we expect to be able to circumvent this problem in large part. 

The next section presents the results of experiments that show that the {\em HNG} algorithm is indeed practically effective. 
Also, the next section analyzes results regarding various aspects of {\em HNG}'s behavior in more depth.  

\section{Experiments and Discussion} \label{experiments}

\subsection{General Setup}

We set out to see whether we could use the {\em HNG} method to achieve translation quality improvements by gathering additional translations to add to the training
data of the entire LDC language pack, including its dictionary. In particular, we wanted to see if we could achieve translation improvements on top of already 
state-of-the-art performing systems trained already on the {\em entire} LDC corpus. Note that at the outset this is an ambitious endeavor (recall the flattening of
the curves in Figure~\ref{f:flatteningCurves} from Section~\ref{introduction}).

\newcite{snow2008} explored the use of the Amazon Mechanical Turk (MTurk) web service for gathering annotations for a variety of natural language 
processing tasks and recently MTurk has been shown to be a quick, cost-effective way to gather Urdu-English translations \cite{bloodgood2010}.
We used the MTurk web service to gather our annotations. 
Specifically, we first crawled a large set of BBC articles on the internet in Urdu and used this as our unlabeled pool from which to gather annotations. 
We applied the {\em HNG} method from Section~\ref{method} to determine what to post on MTurk for workers to translate.\footnote{For practical reasons we restricted
ourselves to not considering sentences that were longer than 60 Urdu words, however.}
We gathered 20,580 n-gram translations for which we paid \$0.01 USD per translation, giving us a total cost of \$205.80 USD. 
We also gathered 1632 randomly chosen Urdu sentence translations as a control set, for which we paid \$0.10 USD per sentence translation.\footnote{The prices we paid
were not market-driven. We just chose prices we thought were reasonable. In hindsight, given how much quicker the phrase translations are for people we could have
had a greater disparity in price.}

\subsection{Accounting for Translation Time}

MTurk returns with each assignment the ``WorkTimeInSeconds.'' This is the amount of time between when a worker accepts an assignment and 
when the worker submits the completed assignment. 
We use this value to estimate annotation times.\footnote{It's imperfect because of network delays and if a person is multitasking or pausing between 
their accept and submit times. Nonetheless, the times ought to be better estimates as they are taken over larger samples.}

\begin{figure}[t]
\begin{center}
\includegraphics[width=.5\textwidth]{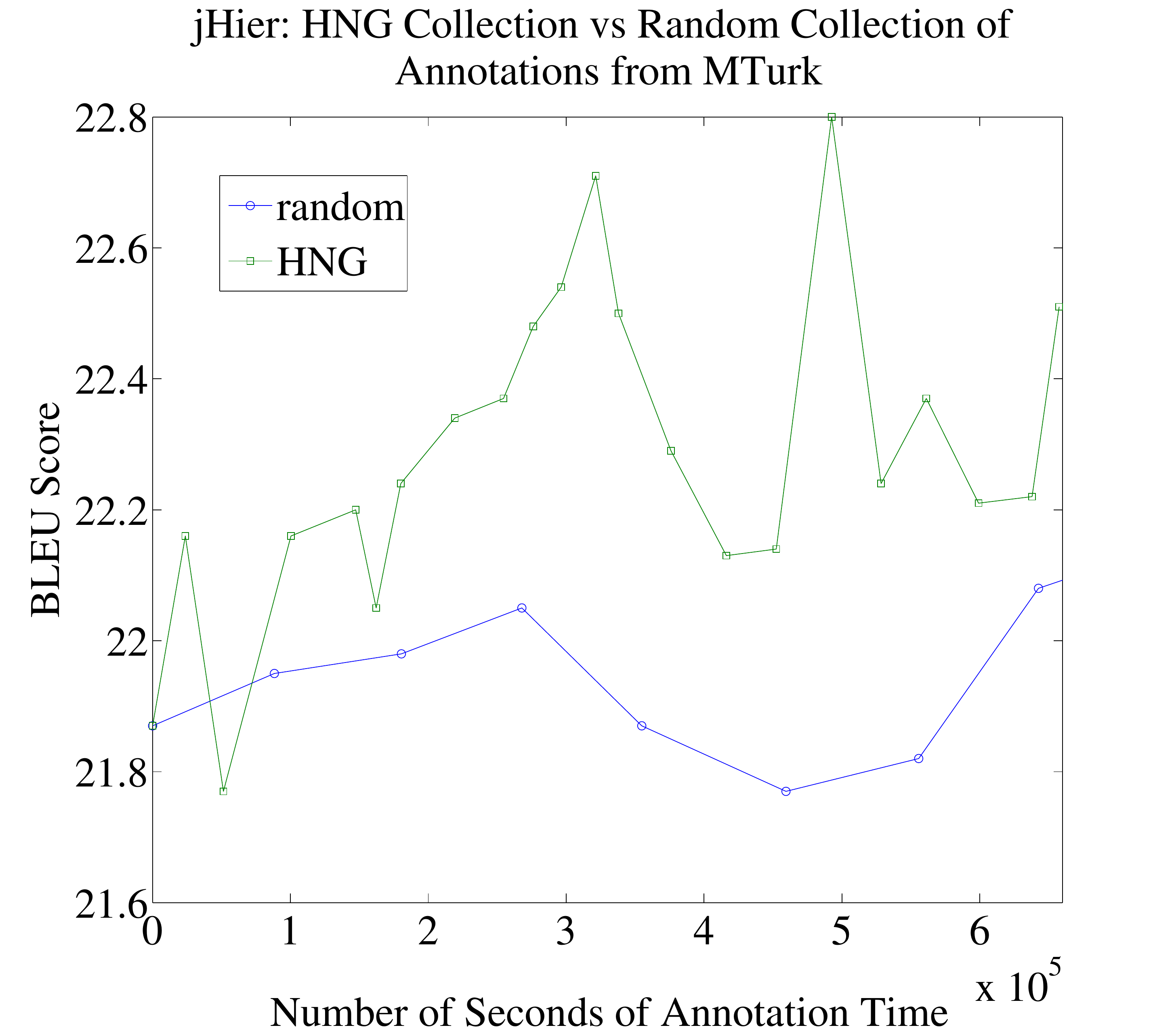}
\caption{\label{f:jHieroHNGRandom}{\em HNG} vs Random collection of new data via MTurk. y-axis measures BLEU. x-axis measures annotation time in seconds.}
\end{center}
\end{figure}

\begin{figure}[t]
\begin{center}
\includegraphics[width=.5\textwidth]{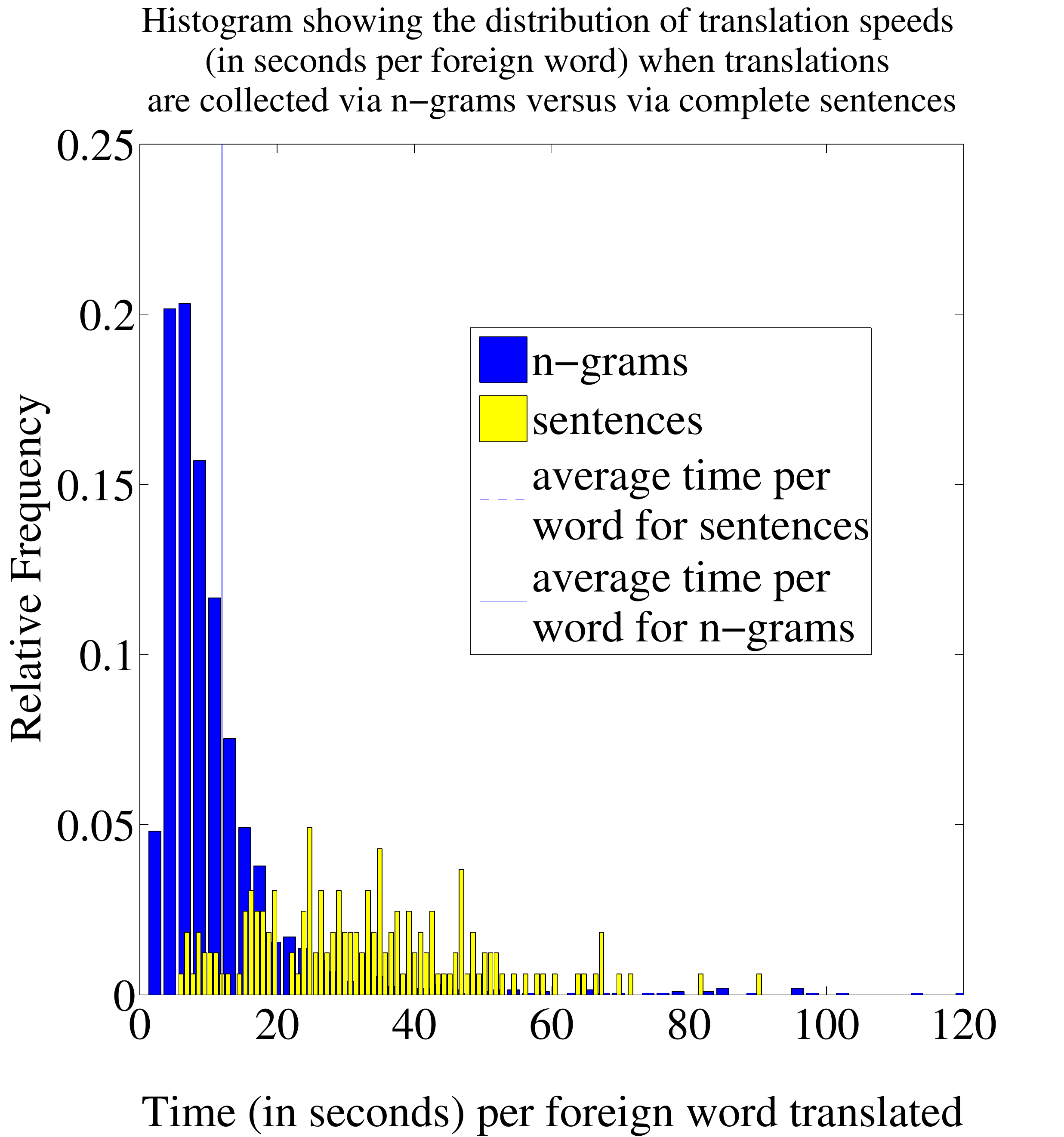}
\caption{\label{f:histogram}Distribution of translation speeds (in seconds per word) for {\em HNG} postings versus complete sentence postings. 
The y-axis measures relative frequency. The x-axis measures translation speed in seconds per word (so farther to the left is faster).}
\end{center}
\end{figure}

Figure~\ref{f:jHieroHNGRandom} shows {\em HNG} collection versus random collection from MTurk. 
The x-axis measures the number of seconds of annotation time. 
Note that {\em HNG} is more effective. 
A result that may be particularly interesting is that {\em HNG} results in a time speedup by more than just the reduction in translated words 
would indicate. 
The average time to translate a word of Urdu with the sentence postings to MTurk was 32.92 seconds. 
The average time to translate a word with the {\em HNG} postings to MTurk was 11.98 seconds. 
This is nearly three times faster. 
Figure~\ref{f:histogram} shows the distribution of speeds (in seconds per word) for {\em HNG} postings versus complete sentence postings. 
Note that the {\em HNG} postings consistently result in faster translation speeds than the sentence postings\footnote{The average speed for the {\em HNG} 
postings seems to be slower than the histogram indicates. This is because there were a few extremely slow outlier
speeds for a handful of {\em HNG} postings. These are almost certainly not cases when the turker is working continuously on the task and so the 
average speed we computed for the {\em HNG} postings might be slower than the actual speed and hence the true speedup may even be faster than 
indicated by the difference between the average speeds we reported.}. 

We hypothesize that this speedup comes about because when translating a full sentence, there's the time required to examine each word and translate them 
in some sense (even if not one-to-one) and then there is an extra significant overhead time to put it all together and synthesize into a larger 
sentence translation. 
The factor of three speedup is evidence that this overhead is significant effort compared to just quickly translating short n-grams from a sentence. 
This speedup is an additional benefit of the {\em HNG} approach. 

\subsection{Bucking the Trend}

\begin{figure}[t]
\begin{center}
\includegraphics[width=.5\textwidth]{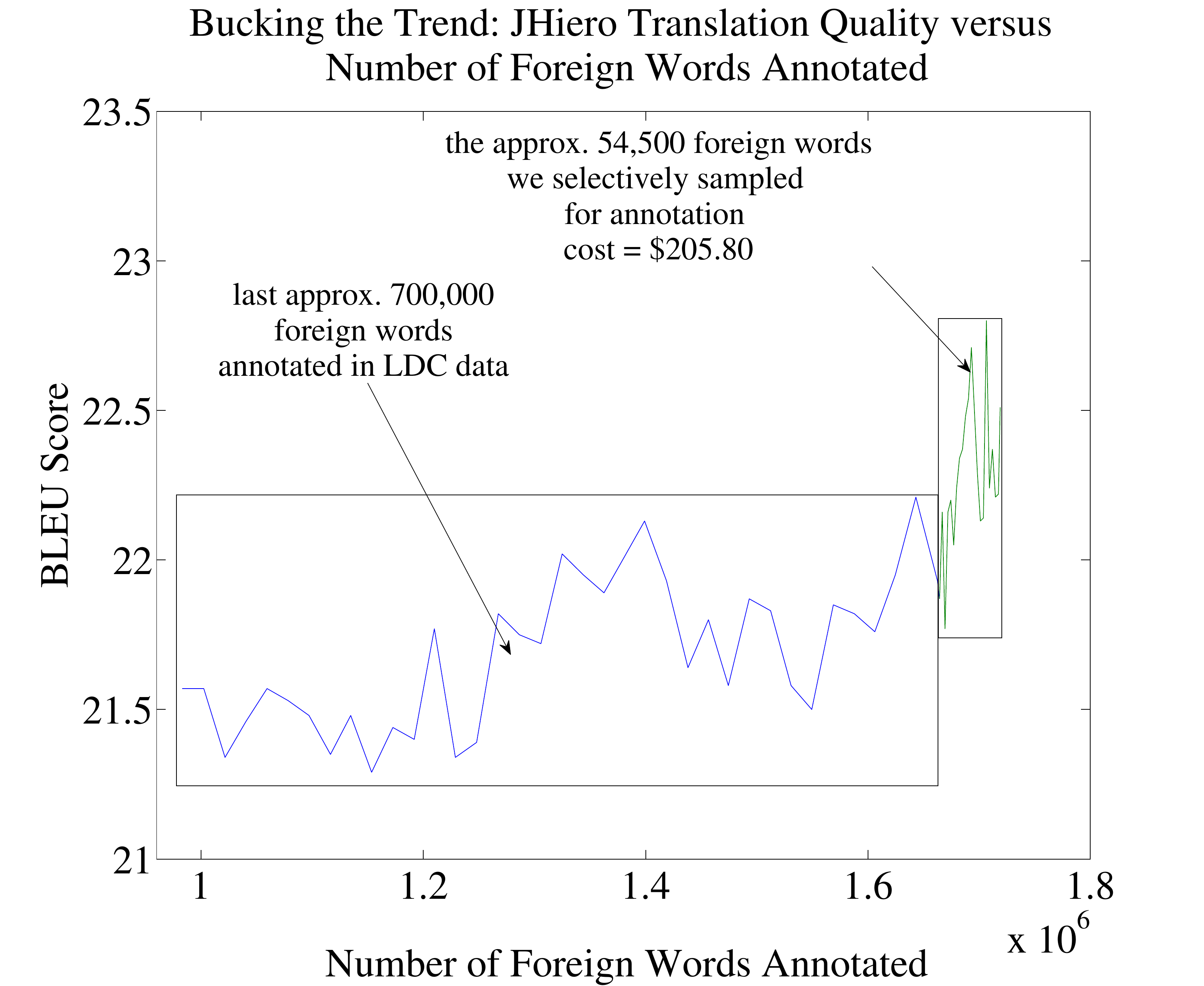}
\caption{\label{f:jHieroBucking}Bucking the trend: performance of {\em HNG}-selected additional data from BBC web crawl data annotated via Amazon Mechanical Turk.
y-axis measures BLEU. x-axis measures number of words annotated.}
\end{center}
\end{figure}

We gathered translations for $\approx$ 54,500 Urdu words via the use of {\em HNG} on MTurk. 
This is a relatively small amount, $\approx$ 3\% of the LDC corpus.
Figure~\ref{f:jHieroBucking} shows the performance when we add this training data to the LDC corpus. 
The rectangle around the last 700,000 words of the LDC data is wide and short (it has a height of 0.9 BLEU points and a width of 700,000 words) but the rectangle around the 
newly added translations is narrow and tall (a height of 1 BLEU point and a width of 54,500 words). 
Visually, it appears we are succeeding in bucking the trend of diminishing returns. 
We further confirmed this by running a least-squares linear regression on the points of the last 700,000 words annotated in the LDC data and also for the points in
the new data that we acquired via MTurk for \$205.80 USD. We find that the slope fit to our new data is 6.6245E-06 BLEU points per Urdu word, or 6.6245 BLEU points
for a million Urdu words. The slope fit to the LDC data is only 7.4957E-07 BLEU points per word, or only 0.74957 BLEU points for a million words. 
This is already an order of magnitude difference that would make the difference between it being worth adding more data and not being worth it; and this is leaving
aside the added time speedup that our method enjoys.

Still, we wondered why we could not have raised BLEU scores even faster. 
The main hurdle seems to be one of coverage. 
Of the 20,580 n-grams we collected, only 571 (i.e., 2.77\%) of them ever even occur in the test set. 

\subsection{Beyond BLEU Scores}

\begin{figure}
\begin{center}
\includegraphics[width=.45\textwidth]{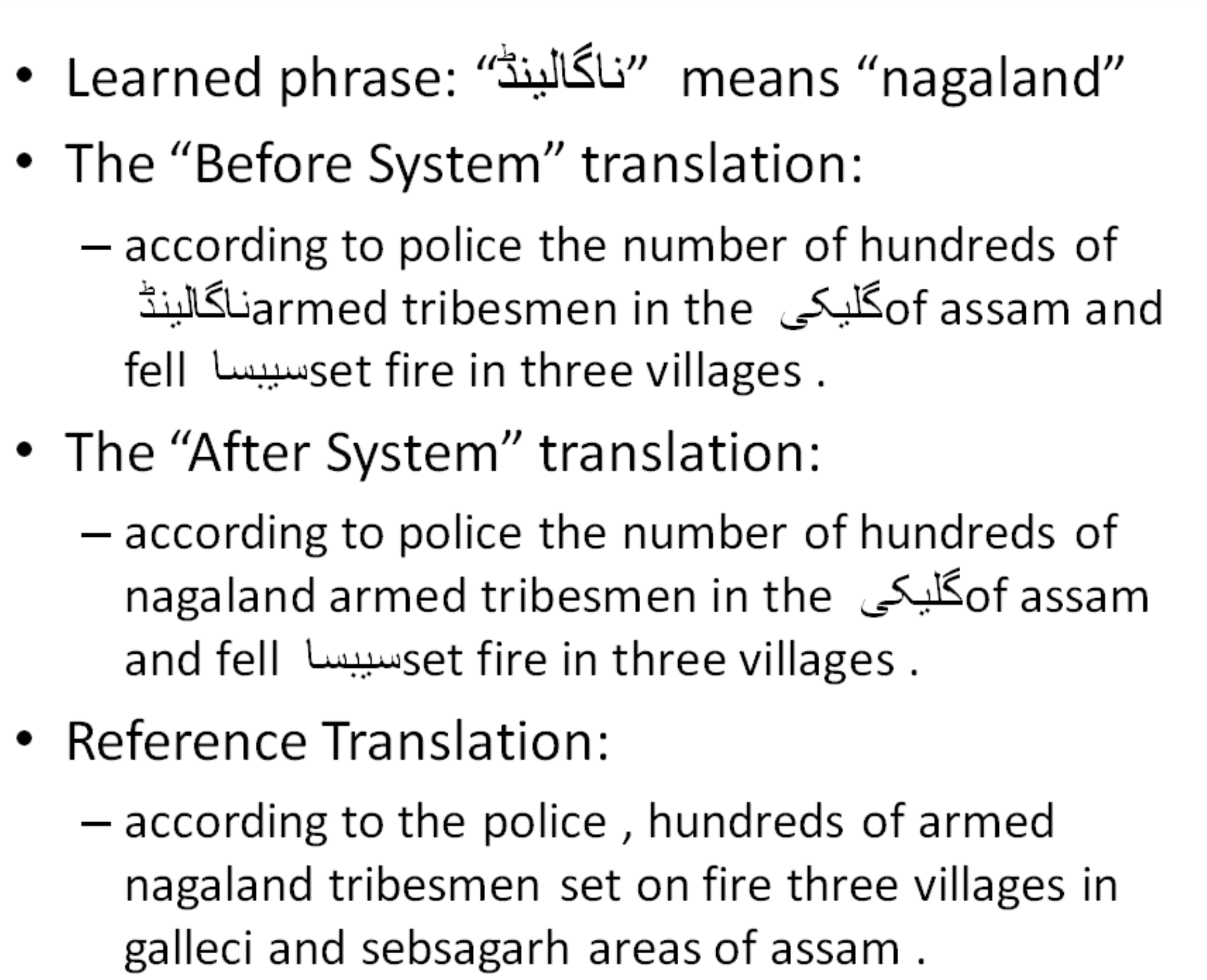}
\caption{\label{f:nagaland}Example of strategy working.}
\end{center}
\end{figure}

BLEU is an imperfect metric \cite{callison-burch2006}. One reason is that it rates all ngram mismatches equally although some are much more important 
than others. Another reason is it's not intuitive what a gain of x BLEU points means in practice. 
Here we show some concrete example translations to show the types of improvements we're achieving and also some 
examples which suggest improvements we can make to our AL selection algorithm in the future.  
Figure~\ref{f:nagaland} shows a prototypical example of our system working. 

\begin{figure}
\begin{center}
\includegraphics[width=.45\textwidth]{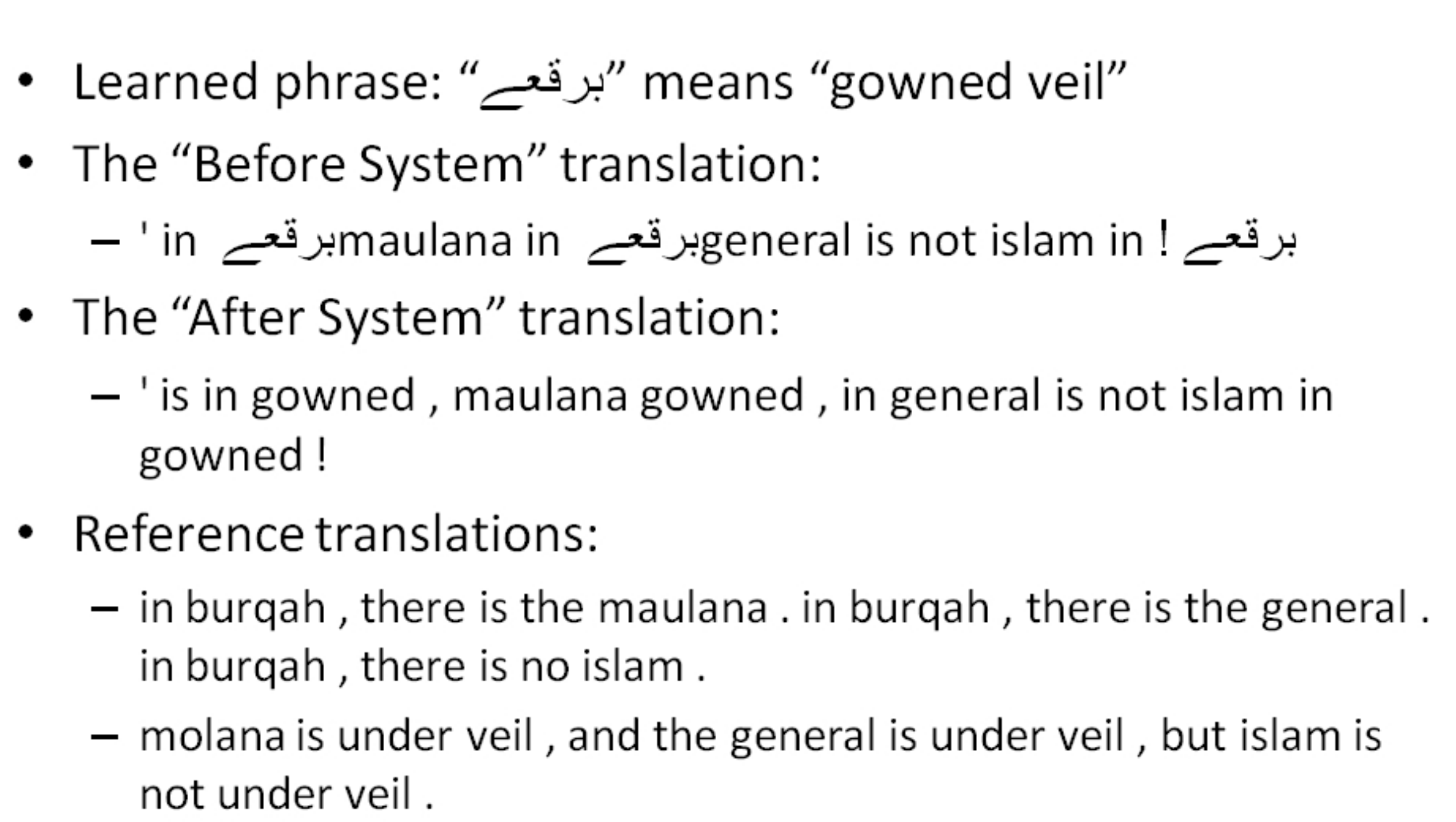}
\caption{\label{f:gownedVeilExample}Example showing where we can improve our selection strategy.}
\end{center}
\end{figure}

Figure~\ref{f:gownedVeilExample} shows an example where the strategy is working partially but not as well as it might.
The Urdu phrase was translated by turkers as ``gowned veil''. 
However, since the word aligner just aligns the word to ``gowned'', we only see
``gowned'' in our output.
This prompts a number of discussion points. First, the `after system' has better translations but they're not rewarded by BLEU scores because the references use the words
`burqah' or just `veil' without `gowned'. Second, we hypothesize that we may be able to see improvements by overriding the automatic alignment software whenever we obtain a
many-to-one or one-to-many (in terms of words) translation for one of our trigger phrases. In such cases, we'd like to make sure that every word on the `many' side is aligned to the
single word on the `one' side. For example, we would force both `gowned' and `veil' to be aligned to the single Urdu word instead of allowing the automatic aligner to only align
`gowned'. 

\begin{figure}
\begin{center}
\includegraphics[width=.45\textwidth]{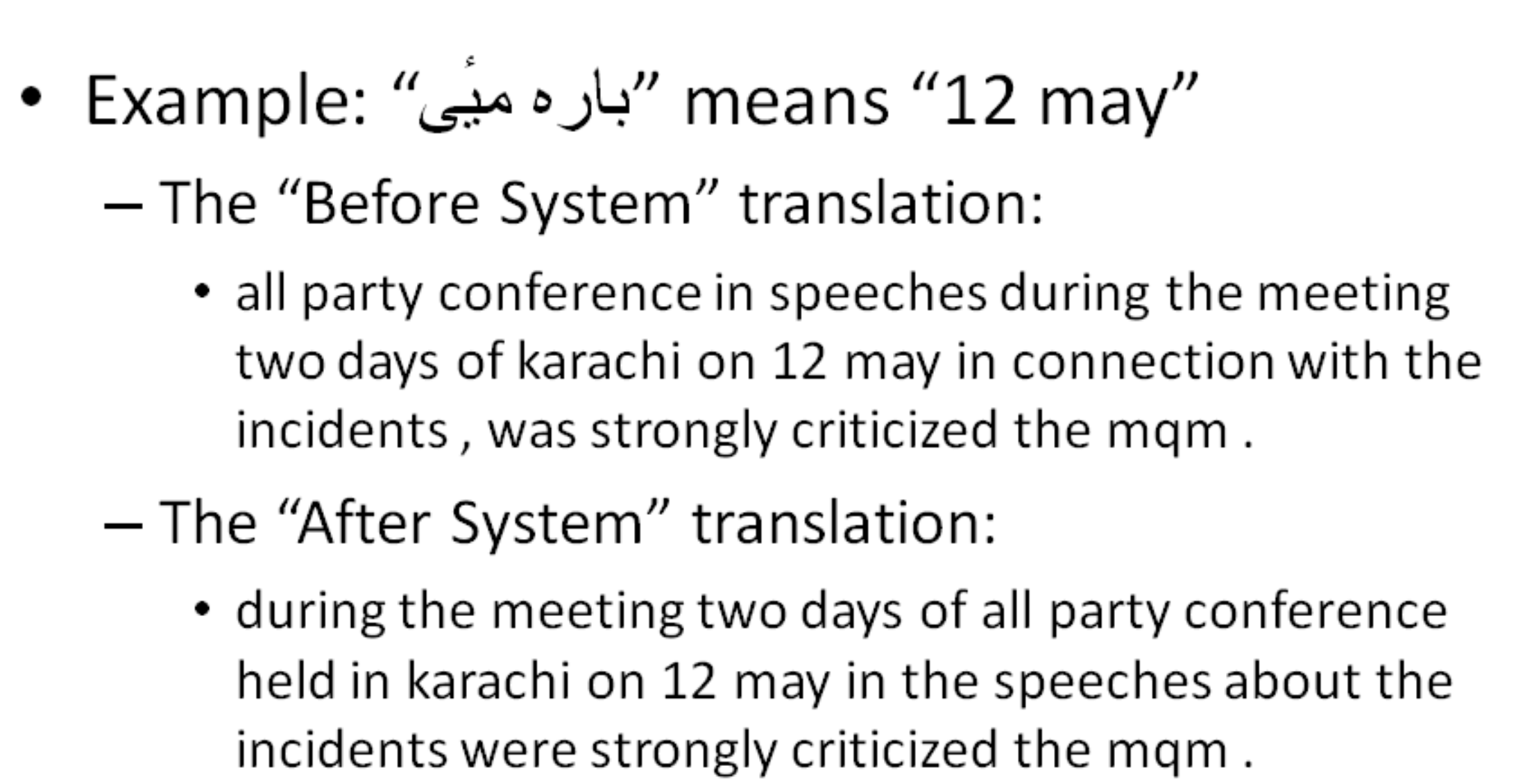}
\caption{\label{f:12MayExample}Example showing where we can improve our selection strategy.}
\end{center}
\end{figure}

Figure~\ref{f:12MayExample} shows an example where our ``before'' system already got the translation correct without the need for the additional phrase translation.
This is because though the ``before'' system had never seen the Urdu expression for ``12 May'', it had seen the Urdu words for ``12'' and ``May'' in isolation and was able to successfully compose them.
An area of future work is to use the ``before'' system to determine such cases automatically and avoid asking humans to provide translations in such cases. 

\section{Conclusions and Future Work} \label{conclusions}

We succeeded in bucking the trend of diminishing returns and improving translation quality while keeping annotation costs low. 
In future work we would like to apply these ideas to domain adaptation (say, general-purpose MT system to work for scientific domain such as chemistry). 
Also, we would like to test with more languages, increase the amount of data we can gather, and investigate stopping criteria further. 
Also, we would like to investigate increasing the efficiency of the selection algorithm by addressing issues such as the one raised by the 12 May example presented
earlier. 

\section*{Acknowledgements}
This work was supported by the Johns Hopkins University Human Language
Technology Center of Excellence. Any opinions, findings,
conclusions, or recommendations expressed in this material are those of the authors and do not necessarily reflect
the views of the sponsor.

\bibliographystyle{acl}

\bibliography{paper}

\end{document}